\documentclass[final]{cvpr}

\usepackage{times}
\usepackage{epsfig}
\usepackage{graphicx}
\usepackage{amsmath}
\usepackage{amssymb}
\usepackage{graphicx}
\usepackage{tabularx}
\usepackage{amsthm}
\newcolumntype{Y}{>{\centering\arraybackslash}X}
\usepackage{booktabs}
\usepackage{multirow}
\usepackage{float} 
\usepackage{subfigure}
\usepackage[ruled,linesnumbered]{algorithm2e}

\usepackage[pagebackref=true,breaklinks=true,colorlinks,bookmarks=false]{hyperref}

\def\cvprPaperID{4743} 
\def\confYear{CVPR 2021}

\begin{document}

\title{Bag of Tricks and A Strong Baseline for Image Copy Detection}

\author{Wenhao Wang, Weipu Zhang, Yifan Sun, Yi Yang\\
Baidu Research\\
{\tt\small wangwenhao0716@gmail.com, zhangweipu01@baidu.com, sunyifan01@baidu.com, yee.i.yang@gmail.com}
}

\maketitle

\begin{abstract}
\vspace*{-2mm}
Image copy detection is of great importance in real-life social media. In this paper, a bag of tricks and a strong baseline are proposed for image copy detection. Unsupervised pre-training substitutes the commonly-used supervised one. Beyond that, we design a descriptor stretching strategy to stabilize the scores of different queries. Experiments demonstrate that the proposed method is effective. The proposed baseline ranks third out of $526$ participants on the Facebook AI Image Similarity Challenge: Descriptor Track. The code and trained models are available \href{https://github.com/WangWenhao0716/ISC-Track2-Submission}{here}.
\vspace*{-4mm}

\end{abstract}

\section{Introduction}

The goal of image copy detection is to determine whether a query image is a modified copy of any images in a reference dataset. It has extensive numbers of applications, such as checking integrity-related problems in social media. Although this topic has been researched for decades, and it has been deemed as a solved problem, most state-of-the-art solutions cannot deliver satisfying results under real-life scenarios \cite{douze20212021}. There are two main reasons. First, real-life cases in social media involve billions to trillions of images, which introduces many ``distractor” images to degrade the performance. Second, the transformations used to edit images are countless. It is very challenging for an algorithm to be robust to unseen scenarios.\par

In this paper, we provide a bag of tricks and a strong baseline to compete for Image Similarity Challenge: Descriptor Track at NeurIPS'21 (ISC2021) \cite{douze20212021}. This competition builds a benchmark that features a variety of image transformations to mimic real-life cases in social media. To mimic a needle-in-haystack setting, both the query and reference set contain a majority of ``distractor" images that do not match. The evaluation metric adopted is micro Average Precision, which penalizes any detected pair for a distractor query.\par
\begin{figure}[t]
\centering 
\includegraphics[width=0.47\textwidth]{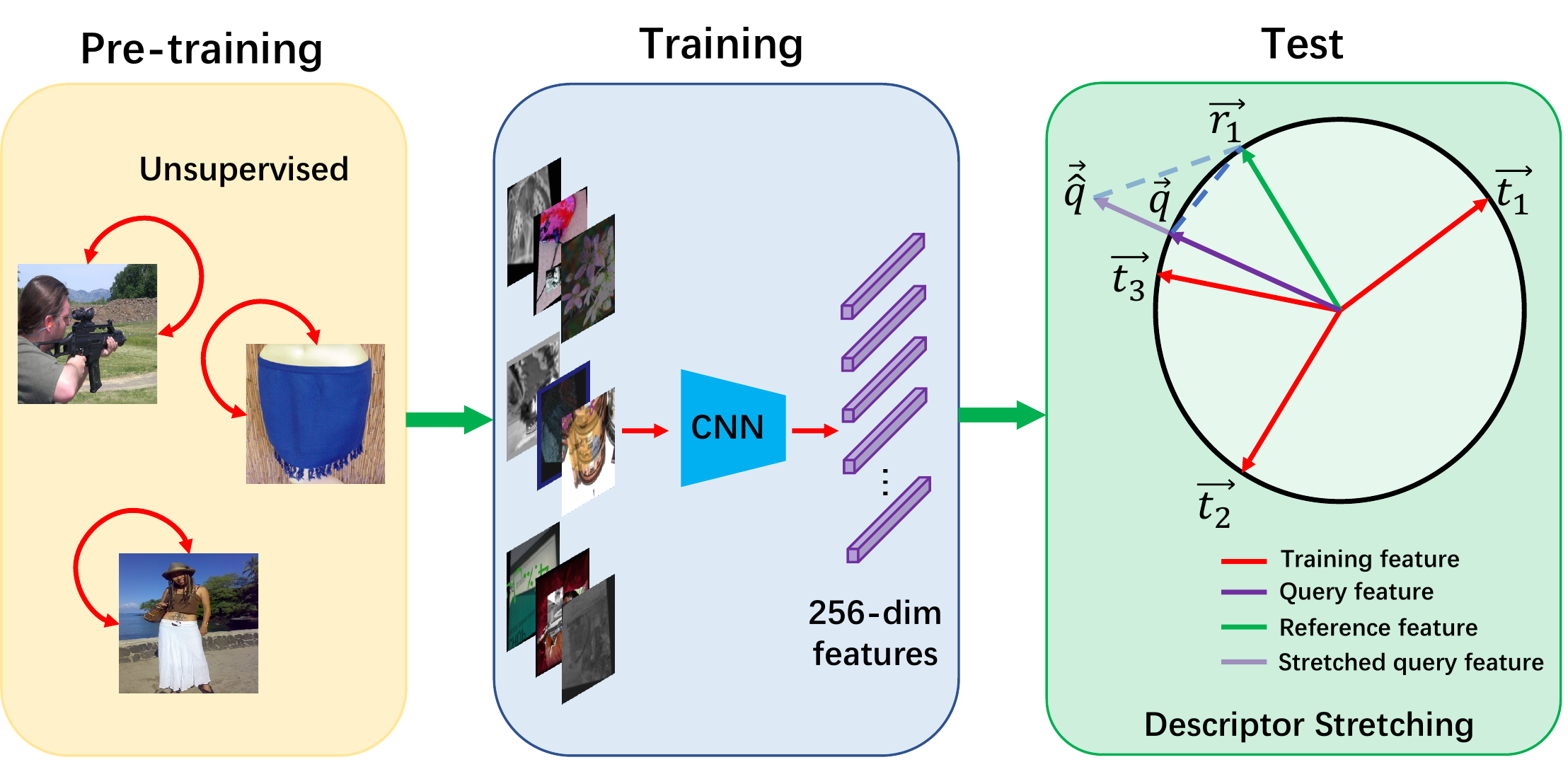}
\vspace*{2mm}
\caption{The designed strong baseline. Unsupervised pre-training is used to substitute supervised one. The strong baseline is trained with a bag of tricks on augmented images. Descriptor stretching, which stabilizes the scores of different queries, is proposed for testing.} 
\label{frame_1}
\end{figure}

The approach consists of three parts, \ie pre-training, training, and test. In pre-training, unsupervised pre-training on ImageNet \cite{deng2009imagenet} instead of the commonly-used supervised pre-training is performed. Specifically, we empirically find that Barlow-Twins \cite{Zbontar2021BarlowTS} pre-training is superior to some other unsupervised pre-training methods. In training, a strong deep metric learning baseline is designed by combining a bag of tricks. Moreover, to make the sample pairs more informative, some image augmentations are employed to generate training images. The diversity of augmentations promotes learning robust representation. During testing, the Descriptor Track setting directly uses the Euclidean distance as the final matching score and thus prohibits the score normalization operation in Matching Track. In response to this characteristic, we propose a novel descriptor stretching strategy. It rescales (stretches) the descriptors to stabilize the scores of different queries. In another word, it approximates the score normalization effect by directly stretching the descriptors. Empirically, this stretching strategy significantly improves retrieval accuracy. The illustration of the proposed approach is shown in Fig. \ref{frame_1}. \par 
\begin{figure*}[t]
\centering 
\includegraphics[width=0.99\textwidth]{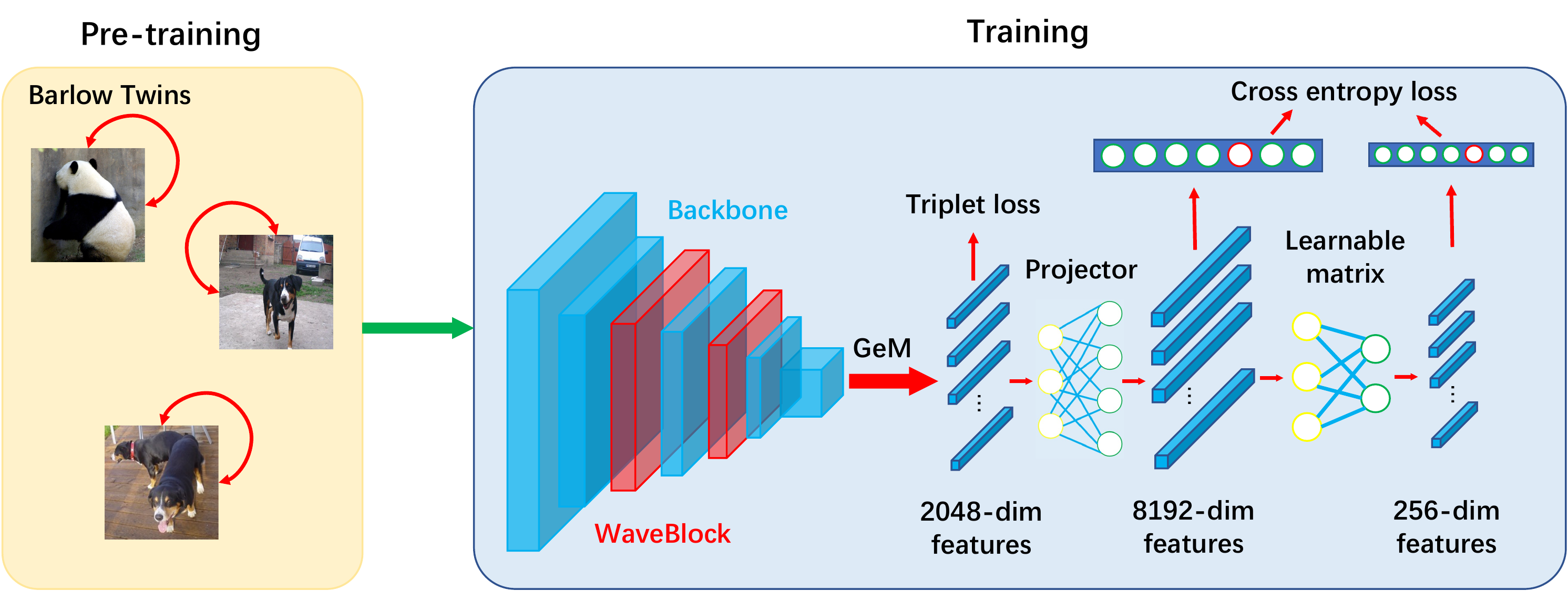}
\vspace*{2mm}
\caption{The bag of tricks and the strong baseline. A recent self-supervised learning method, Barlow-Twins \cite{Zbontar2021BarlowTS}, is performed for pre-training. During training, the built baseline includes GeM \cite{radenovic2018fine}, WaveBlock \cite{wang2020attentive}, a high-dimension projector, a learnable matrix, two commonly-used losses, and a warming up with cosine annealing learning rate. } 
\label{frame_2}
\end{figure*}
In summary, the main contributions of this paper are:
\begin{enumerate}
 \item The paper provides a bag of tricks and a strong baseline for image copy detection.
  \item The proposed baseline handles real-life copy detection scenarios in social media well.
 \item Our baseline ranks third out of $526$ participants in the Image Similarity Challenge: Descriptor Track at NeurIPS’21.
\end{enumerate}

\section{Related Work}

\subsection{Copy Detection}
Although copy detection plays an important role in social media, the publication of copy detection is not well known because organizations want to keep copy detection techniques obscure, and the researchers often consider that the task is easy \cite{douze20212021}. For classical approaches, some methods are used to extract global \cite{kim2003content, wan2008survey, hsiao2007new} and local descriptors \cite{amsaleg2001content, berrani2003robust, ke2004efficient}. For deep learning methods, a descriptor, or known as a feature, is extracted by convolutional neural networks \cite{liu2020content, zhou2020cnn}. However, none of them gives a satisfactory performance on challenging, large benchmarks like ISC2021.
\subsection{Unsupervised Pre-training}
The unsupervised pre-trained models are from recent self-supervised learning methods. They are trained to extract image embedding unsupervisedly. MoCo \cite{he2020momentum} builds a dynamic dictionary with a queue and a moving-averaged encoder to conduct contrastive learning. BYOL \cite{jean-bastien2020bootstrap} and its Momentum$^2$ Teacher version  achieve a new state-of-the-art without negative pairs. Other self-supervised methods, such as Barlow Twins \cite{Zbontar2021BarlowTS}, SimSiam \cite{chen2021exploring}, and SwAV \cite{caron2020unsupervised}, also show promising performance. \par 
For copy detection in ISC2021, we find that unsupervised pre-trained models show superior performance to their fully-supervised counterparts. That may be because the category defined in unsupervised pre-training is much more similar to that in copy detection than in fully-supervised one.
\begin{figure*}[t]
\centering 
\includegraphics[width=1\textwidth]{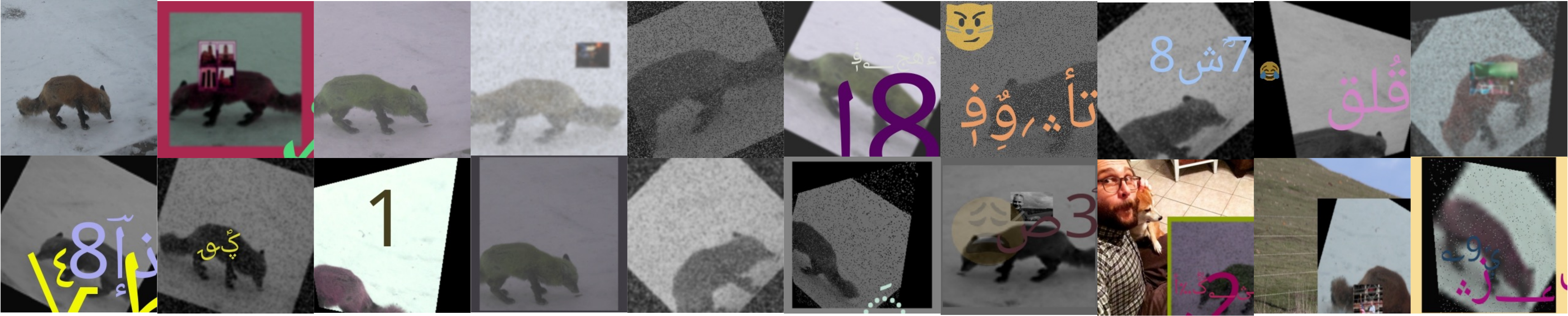}
\vspace*{2mm}
\caption{The set of basic augmentations. It includes random resized cropping, random rotation, random pixelization, random pixels shuffling, random perspective transformation, random padding, random image underlay, random color jitter, random blurring, random grayscale, random horizontal flipping, random Emoji overlay, random text overlay, random image overlay, and resizing. The first image is the resized original image.} 
\label{basic}
\end{figure*}
\subsection{Deep Metric Learning}
Learning a discriminative feature is a crucial component for many tasks, such as image retrieval \cite{qian2019softtriple, kim2020proxy}, face verification \cite{Sun_2020_CVPR, wang2017normface}, and object re-identification \cite{sun2018beyond, zheng2020vehiclenet}. The commonly-used loss functions can be divided into two classes: pair-based \cite{hermans2017defense, sohn2016improved, oh2016deep} and proxy-based \cite{Liu_2017_CVPR, wang2018cosface, deng2019arcface} losses. Circle loss \cite{Sun_2020_CVPR} gives a unified formula for two paradigms. \par 
In our solution to ISC2021, we use two common losses, \ie triplet loss with hard sample mining and cross-entropy loss. The two losses are proved to be simple but effective.
\section{Proposed Method}
In this section, we introduce each important component in the proposed baseline. In the training part, we discuss the tricks to train the baseline and the designed augmentations. In the test part, we discuss the proposed descriptor stretching. The whole baseline is shown in Fig. \ref{frame_2}.

\subsection{Unsupervised Pre-training}
In the ISC2021, a category is defined at a very ``tight" level, \ie images generated by exact duplication, near-exact duplication, and edited copy are considered in the same category while two images that belong to the same instance or object are not in the same category. However, when using fully-supervised pre-training on the ImageNet, two images share the same instance or object are in one class. Therefore, the definition of the category is contradicted between the ISC2021 and ImageNet. \par
Fortunately, the recent research on self-supervised learning provides a new direction. It defines every image as a category and uses invariance to data augmentation as their main training criterion \cite{douze20212021}. Although it seems natural to directly adopt a self-supervised learning method to train a model for copy detection, we choose to give up this solution for two reasons. First, the self-supervised learning methods often cost a lot of days to get convergence results. For instance, on ImageNet, BYOL \cite{jean-bastien2020bootstrap} or its Momentum$^2$ Teacher version \cite{li2021momentum} takes about two weeks to pre-train $300$ epochs with ResNet-152 \cite{he2016deep} backbone using $8$ NVIDIA Tesla V100 32GB GPUs. In addition, when using sophisticated transformations online, the time becomes much longer, which is unaffordable. Second, even if the resource is unlimited, we do not get a satisfying performance only trained by self-supervised learning methods, which may be because that the hyper-parameters are not adjusted carefully.\par
As a result, we choose to use Barlow-Twins \cite{Zbontar2021BarlowTS} to get the pre-trained model on ImageNet. The used augmentations follow the default setting in the original implementation.

\subsection{Training Tricks}
The original version of the baseline is from DomainMix \cite{wang2021domainmix}. To be more powerful, we introduce some recent methods and adapt them to the Image Similarity Challenge: Descriptor Track.\par
\textbf{Generalized mean pooling.} Generalized mean (GeM) pooling \cite{radenovic2018fine} is proposed to compute the  generalized mean of a three-dim tensor: 
\begin{equation}
\mathbf{f}^{(g)}=\left[\mathrm{f}_{1}^{(g)} \ldots \mathrm{f}_{k}^{(g)} \ldots \mathrm{f}_{K}^{(g)}\right]^{\top},
\end{equation}
\begin{equation}
\mathrm{f}_{k}^{(g)}=\left(\frac{1}{\left|\mathcal{X}_{k}\right|} \sum_{x \in \mathcal{X}_{k}} x^{p_{k}}\right)^{\frac{1}{p_{k}}},
\end{equation}
where: $p_{k}$ is a learnable pooling parameter. The GeM pooling degenerates to maxing pooling when $p_{k} \rightarrow \infty$, and mean pooling when $p_{k}=1$. This can be easily proved by mathematical analysis. We experimentally set the initialization of the pooling parameter as $3$. The network focuses on the salient area with a larger $p_{k}$, while it learns globally with a smaller $p_{k}$. For image copy detection, we find that $p_{k}$ tends to increase, which shows that the matching of salient areas matters.

\par
\textbf{WaveBlock.} The WaveBlock is proposed in AWB \cite{wang2020attentive} to deal with the noise in pseudo-labels. We further discover WaveBlock can be regarded as a kind of augmentation method at the feature level. Following the original paper, the WaveBlocks are arranged after Stage $2$ and Stage $3$ when using ResNet50 \cite{he2016deep}. The same hyper-parameters are also used. We argue that WaveBlocks enhance the model ability for unseen augmentations. \par
\textbf{High-dimension projector.} We design a projector to project a learned 2048-dim feature into 8192-dim via some linear and non-linear connections. Specifically, the projector adopts the structure of \texttt{FC-BN-LeakyReLU-FC} where the input dimension, hidden dimension, and output dimension are set to $2048$, $4096$, and $8192$, respectively. Empirically, we find that when the IDs in the training set are small, learning from high dimension space is not very useful. However, when the IDs become larger, such as $100,000$, the projector makes a difference. \par  
\textbf{Learnable Matrix.} Per the official rule for Track 2, the maximum dimension of a descriptor is $256$. Therefore, an $8192 \times 256$ matrix is learned during training. By using this matrix, the learned $8192$-dim features can be transformed to $256$-dim ones. This matrix is initialized randomly.\par 
\textbf{Losses.} In this baseline, we use two commonly-used losses, \ie triplet loss with hard sample mining and cross-entropy loss. The cross-entropy loss is used three times. First, it is used for the classification of $8192$-dim features. Second, it is used for the classification of $256$-dim features. Third, we also design a soft cross-entropy loss. The prediction results of $8192$-dim features are regarded as labels to conduct the learning process of $256$-dim ones. The $2048$-dim feature is used for triplet loss with hard sample mining. \par 
\textbf{Warming up with cosine annealing learning rate.} The adopted learning rate is with warming up scheduling and cosine annealing. The basic learning rate is $3.5 \times 10^{-4}$. We train the baseline for $25$ epochs. The ratio for scheduling is 
\begin{equation}
ratio = \left\{ \begin{array}{l}
0.99 \cdot epoch/5 + 0.01,0 \le epoch < 5\\
1,5 \le epoch < 10\\
0.5 \cdot \left( {\cos \left( {{\textstyle{{epoch - 10} \over {25 - 10}}} \cdot \pi } \right) + 1} \right),10 \le epoch < 25
\end{array} \right.
\end{equation}
\par


\subsection{Augmentation}
We design a set of augmentations to transform images. The set of basic augmentations includes random resized cropping, random rotation, random pixelization, random pixels shuffling, random perspective transformation, random padding, random image underlay, random color jitter, random blurring, random grayscale, random horizontal flipping, random Emoji overlay, random text overlay, random image overlay, and resizing. A display of using all the basic augmentations on one image is shown in Fig. \ref{basic}. The first image is the resized original image. \par

\subsection{Descriptor Stretching}
\begin{figure}[t]
\centering 
\includegraphics[width=0.44\textwidth]{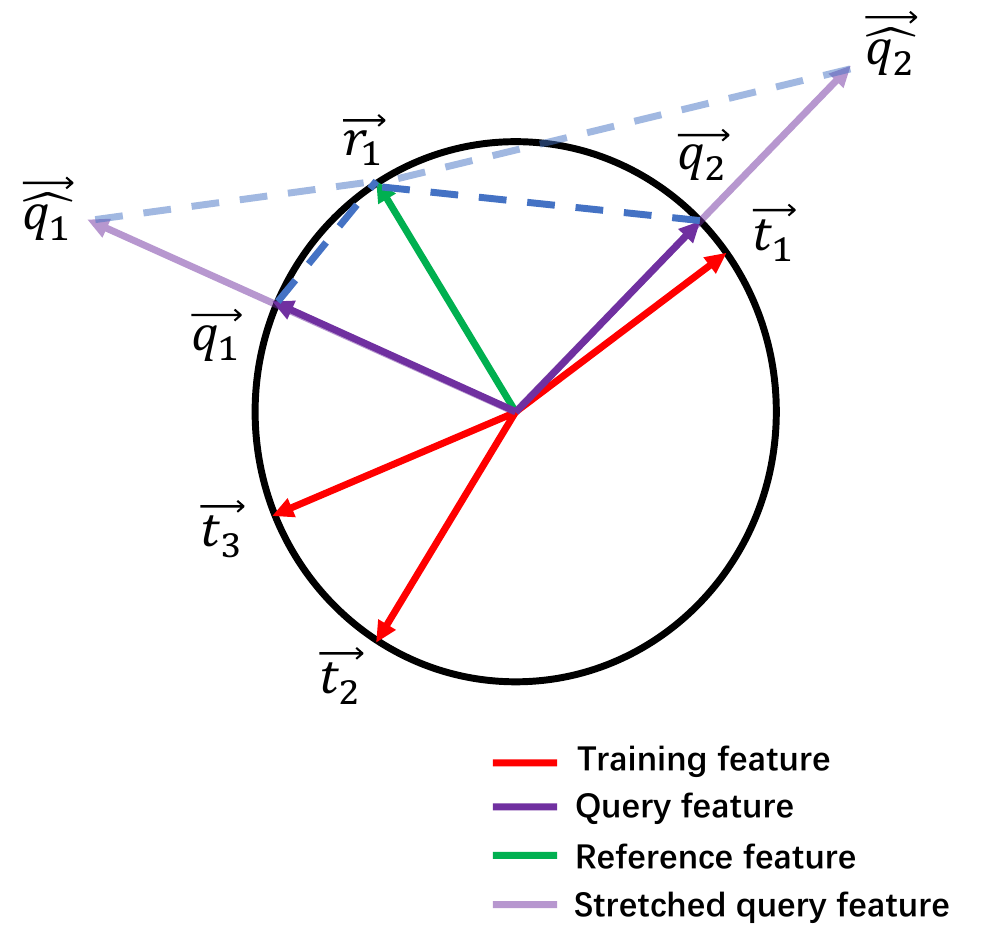}
\vspace*{2mm}
\caption{The illustration of the proposed descriptor stretching strategy. The $L_2$ norms of all the features except for the stretched ones are all equal to $1$.}
\label{ds}
\end{figure}
In the Descriptor Track, the Euclidean distance is used as a matching score directly while the scores of different queries are not comparable. In Matching Track, this phenomenon also exists, and thus the score normalization \cite{jegou2011exploiting} is used. However, in the Descriptor Track, score normalization, which applies to similarity scores, is no longer appropriate. To stabilize the scores of different queries, we propose a descriptor stretching strategy. The illustration of the strategy is shown in Fig. \ref{ds}.\par 
Given the feature of a query image $\vec{q_1}$, and a reference image $\vec{r_1}$, the original score $s_{1}$ is defined as 
\begin{equation}
s_{1}=\left| \vec{q_{1}} -\vec{r_{1}} \right|.  
\end{equation}
Similarly, we have:
\begin{equation}
s_{2}=\left| \vec{q_{2}} -\vec{r_{1}} \right|.  
\end{equation}
If $s_{1}>s_{2}$, $\vec{q_{2}}$ is more similar to $\vec{r_{1}}$ than $\vec{q_{1}}$, and vice versa. The definition of descriptor stretching is
\begin{equation}
\vec{\hat{q_{1}}}=\alpha \cdot s_{n_1} \cdot \vec{q_{1}},
\end{equation}
where: $\alpha$ is a hyper-parameter, and $s_{n_1}$ is the mean of the top $n$ inner product scores between $\vec{q_{1}}$ and the features of images from the training set. Then the stretched score $\hat{s_{1}}$ is defined as:
\begin{equation}
\hat{s_{1}}=\left| \vec{\hat{q_{1}}} -\vec{r_{1}} \right|. 
\end{equation}
Similarly, we have:
\begin{equation}
\vec{\hat{q_{2}}}=\alpha \cdot s_{n_2} \cdot \vec{q_{2}},
\end{equation}
\begin{equation}
\hat{s_{2}}=\left| \vec{\hat{q_{2}}} -\vec{r_{1}} \right|. 
\end{equation}
Therefore, after stretching, if $\hat{s_{1}}>\hat{s_{2}}$, $\vec{\hat{q_{2}}}$ is more similar to $\vec{\hat{r_{1}}}$ than $\vec{\hat{q_{1}}}$, and vice versa. We use the stretched feature of a query image as its final descriptor.\par
Empirically, $\alpha$ equals $2.5$, and $n$ equals $5$. Although from theory, it is still unclear that why the descriptor stretching strategy works so well on stabilizing the scores of different queries, its outstanding performance is crucial to our final ranking.

\section{Experiments}
\section{Experimental Settings}
In the unsupervised pre-training part, we choose Barlow-Twins \cite{Zbontar2021BarlowTS}, and the selected backbone is ResNet50 \cite{he2016deep}. We do not re-train the model, and just use the model supplied by the official implement. \par
In training, we select $100,000$ out of $1000,000$ images from the official training set. The criterion is selecting one every $10$. Each image is augmented $19$ times, and thus for training, each ID has $20$ images. The training process takes less than one day on $4$ NVIDIA Tesla V100 32GB GPUs. Each training batch includes $128$ images of $32$ IDs. Adam optimizer \cite{DBLP:journals/corr/KingmaB14} is used to optimize the networks. The image size is $256 \times 256$. The model is trained for $25$ epochs, and in each epoch, we have $8000$ iterations. The basic learning rate is set to $3.5 \times 10^{-4}$.\par
During testing, we use YOLOv5 \cite{glenn_jocher_2020_4154370} to detect overlay automatically. The detected overlays substitute the original images. For training YOLOv5 \cite{glenn_jocher_2020_4154370}, we just generate some images with overlay augmentations and corresponding bounding boxes automatically from the training dataset. We test the trained models for four scales, \ie $200 \times 200$, $256 \times 256$, $320 \times 320$, and $400 \times 400$. To fuse these multi-scale features, we firstly normalize them such that their $L_2$ norm equals $1$, then the normalized features are averaged, and finally, an $L_2$ normalization is applied to produce the final descriptor \cite{yang2021dolg}. Then, we use descriptor stretching to rescale all the fused query features. The fused training and reference features remain the same.\par 

\subsection{Comparison with State-of-the-Arts}
To prove the superiority of the strong baseline, we compare the proposed model with state-of-the-art methods from the leaderboard in Phase $2$. The comparison results are shown in Table \ref{sota}. In the ISC2021, there are $526$ participants, and $21$ teams have submitted their final results. With a bag of tricks, the strong baseline shows promising performance.
\begin{table}
  \caption{Comparison with state-of-the-art methods from the leaderboard in Phase $2$. Recall@Precision $90$ is a secondary metric provided for information purposes only as an indicator of a model's recall at a reasonably good precision level but is not used for ranking purposes. Our results are highlighted in bold.}
  \vspace*{2mm}
\small
  \begin{tabularx}{\hsize}{|p{1.5cm}|YY|}
    \hline
    \multicolumn{1}{|c|}{\multirow{2}{*}{Team}} &
    \multicolumn{2}{c|}{Score}  \\
    \cline{2-3}
      &\footnotesize{Micro-average Precision} & \footnotesize{Recall@Precision 90} \\
    \hline\hline
    lyakaap  & $0.6354$ & $0.6354$ \\
    S-square  & $0.5905$ & $0.5086$ \\
    \textbf{Ours} & $\textbf{0.5788}$ & $\textbf{0.4886}$ \\
    forthedream2  & $0.5736$ & $0.4980$ \\
    Zihao  & $0.5461$ & $0.4813$ \\
    separate  & $0.5312$ & $0.3169$ \\
    \footnotesize{AITechnology}  & $0.5253$ & $0.4191$ \\
    ...&...&... \\
    \hline
    GIST \cite{oliva2001modeling} & $0.0526	$ & $-$ \\ 
    \hline
  \end{tabularx}

  \label{sota}
\end{table}

\subsection{Ablation Studies}
The ablation studies are conducted on $25,000$ query images in Phase $1$ to prove the effectiveness of each component in our method. The evaluation metric is Micro-average Precision. The experimental results are displayed in Table \ref{abla}. \par

\textbf{Comparison between supervised pre-training and unsupervised pre-training.}
First, we discuss the improvement brought by unsupervised pre-training. The experimental results are denoted as ``Supervised" and ``Unsupervised'' in Table \ref{abla}, respectively. We can find that, instead of supervised pre-training, over $14\%$ improvement has been obtained by using unsupervised pre-trained models. This phenomenon may be because the definition of category in unsupervised pre-training is similar to that in the task of copy detection. \par 
\textbf{The improvement from descriptor stretching.} By using descriptor stretching, the Micro-average Precision can be improved by $17.26\%$ significantly. To prove the significant performance indeed from the comparable scores, we also adopt two other metrics, \ie Recall@Rank $1$ and Recall@Rank $10$. The two metrics are calculated according to per query separately. Therefore, the incomparable scores of queries cannot influence it. Before stretching, the two scores are $76.18\%$ and $79.72\%$, respectively. After stretching, the two scores are $76.10\%$ and $79.98\%$, respectively. Therefore, there is no improvement on these two scores, which shows the performance per query is not improved, and the overall performance improvement comes from comparable scores of queries.\par 
\textbf{The improvement from overlay detection and multi-scale testing.} We discover that when image $A$ overlays on another image $B$, image $A$ has a true match in the reference dataset while image $B$ is the background. Therefore, when cropping the image from the background, about $1\%$ improvement can be observed. Multi-scale testing is a common practice in computer vision, and it also shows the effectiveness in the ISC2021. By using multi-scale testing, the performance can be improved from $71.49\%$ to $73.02\%$.\par

\begin{table}
\small
\caption{The ablation study about the proposed baseline. ``Supervised" or ``Unsupervised" denotes the supervised pre-training or unsupervised pre-training is used, respectively. ``$+$ Des-Str'' denotes adding descriptor stretching strategy. ``$+$ Det'' denotes adding overlay detection (YOLOv5 \cite{glenn_jocher_2020_4154370}). ``$+$ Multi'' denotes using multi-scale testing. The best results are highlighted in bold.}
\vspace*{2mm}
  \begin{tabularx}{\hsize}{|p{1.5cm}|YY|}
    \hline
    \multicolumn{1}{|c|}{\multirow{2}{*}{Method}} &
    \multicolumn{2}{c|}{Score}  \\
    \cline{2-3}
      &\footnotesize{Micro-average Precision} & \footnotesize{Recall@Precision 90} \\
    \hline\hline
    Supervised & $0.39089$ & $0.18133$  \\
    Unsupervised& $0.53218$ & $0.29693$   \\
    \hline
    $+$ Des-Str & $0.70481$ & $0.61631$ \\
    $+$ Det & $0.71487$ & $0.62913$ \\
    $+$ Multi & $\textbf{0.73017}$ & $\textbf{0.63975}$ \\
    \hline
  \end{tabularx}

  \label{abla}
\end{table}
\section{Conclusion}
In this paper, we introduce our winning solution to the Image Similarity Challenge at NeurIPS'21. The proposed strong baseline uses recent self-supervised learning methods for pre-training instead of traditional supervised ones. Further, a descriptor stretching strategy is proposed to stabilize the scores of queries. We hope the proposed solution is beneficial for real-life applications including content tracing, copyright infringement, and misinformation.

{\small
\bibliographystyle{ieee_fullname}
\bibliography{egbib}
}

\end{document}